\documentclass[letterpaper, 10 pt, conference]{ieeeconf}  %

\IEEEoverridecommandlockouts                              %

\usepackage{times}
\usepackage{epsfig}
\usepackage{graphicx}
\usepackage{amsmath}
\usepackage{amssymb}
\usepackage[font=small]{caption}
\usepackage{subcaption}
\usepackage[export]{adjustbox}
\usepackage{booktabs}
\usepackage[table]{xcolor}
\usepackage{multirow}
\usepackage{rotating}
\usepackage{tabularx}
\usepackage{gensymb}

\newcommand{\bx}{\mathbf{x}}

\newcommand{\bP}{\mathbf{P}}
\newcommand{\bS}{\mathbf{S}}

\newcommand{\tH}[1]{\textbf{#1}}

\newcommand{\ccw}{\cellcolor{white}}
\newcommand{\txtsub}[2]{$\text{#1}_{\text{#2}}$ }

\definecolor{light-gray}{gray}{0.85}
\colorlet{red}{red!80!gray}
\colorlet{orange}{orange!80!gray}
\colorlet{yellow}{yellow!90!gray}
\colorlet{green}{green!70!gray}
\colorlet{cyan}{cyan!80!gray}
\colorlet{blue}{blue!80!gray}
\colorlet{purple}{purple!80!gray}

\usepackage{tikz}

\newcommand{\defaultmetric}[0]{MeanSSD}

\newif\ifarxiv
\arxivtrue

\newif\ifsplashfig
\splashfigtrue

\makeatletter
\let\NAT@parse\undefined
\makeatother
\usepackage[pagebackref=true,breaklinks=true,letterpaper=true,colorlinks,bookmarks=false]{hyperref}

\title{\LARGE \bf
6-DoF Pose Estimation of Household Objects for Robotic Manipulation:\\  An Accessible Dataset and Benchmark
}

\author{Stephen Tyree, Jonathan Tremblay, Thang To, Jia Cheng, Terry Mosier, Jeffrey Smith, and Stan Birchfield \\
{\tt\small \{styree, jtremblay, thangt, jicheng, tmosier, jeffreys, sbirchfield\}@nvidia.com} \\
NVIDIA
}

\begin{document}

\maketitle
\thispagestyle{empty}
\pagestyle{empty}

\begin{abstract}
We present a new dataset for 6-DoF pose estimation of known objects, with a focus on
robotic manipulation research.
We propose a set of toy grocery objects, whose physical instantiations are readily available for purchase and are appropriately sized for robotic grasping and manipulation.
We provide 3D scanned textured models of these objects, suitable for generating synthetic training data,
as well as RGBD images of the objects in challenging, cluttered scenes
exhibiting partial occlusion, extreme lighting variations, multiple instances per image, and a large variety of poses.
Using semi-automated RGBD-to-model texture correspondences, the images are annotated with ground truth poses 
accurate within a few millimeters.
We also propose a new pose evaluation metric called {ADD-H} based on the Hungarian assignment algorithm that is robust to symmetries in object geometry without requiring their explicit enumeration.
We share pre-trained pose estimators for all the toy grocery objects, along with their baseline performance on both validation and test sets.
We offer this dataset to the community to help connect the efforts of computer vision researchers with the needs of roboticists.\footnote{\url{https://github.com/swtyree/hope-dataset}}
\end{abstract}

\section{Introduction}

Estimating the poses of objects in a scene is important for robotic grasping and manipulation in a variety of domains, such as manufacturing, healthcare, commercial, and households.
In pick-and-place tasks, for example, it is necessary for the robot to know the identities and locations of objects in order to grasp them, to avoid collision with other objects during movement, and to place them relative to other objects.
Yet, even today, it is not uncommon for objects in robotics labs to be modified with fiducial markers such as ARTags~\cite{fiala2005cvpr:artag} or AprilTags~\cite{olson2011icra:apriltags} to support detection and pose estimation.
Such modifications are needed because no widely used, off-the-shelf, general-purpose techniques currently exist.

To bridge the gap between research labs and real-world scenarios, many researchers have developed methods to detect objects and estimate their poses from RGB or RGBD images.
Recent progress on this 6-DoF (``degrees of freedom") pose estimation problem has been significant~\cite{labbe2020,tremblay2018corl:dope,hu2019cvpr:segdriv6d,peng2019cvpr:pvnet,zakharov2019iccv:dpod,li2018eccv:deepim,rad2017iccv:bb8,wang20196-pack,wang2019densefusion,bowen2020iros:se3tracknet,bowen2021iros:bundletrack}.
Although there remain many open research problems, the basics are in place.  

\ifsplashfig
    \begin{figure}
        \centering
        \includegraphics[width=\linewidth]{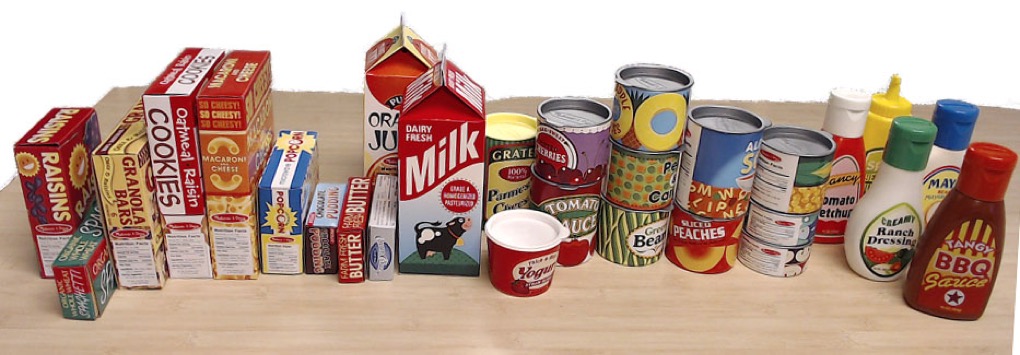}
        \caption{Set of 28 toy grocery objects used in the dataset.}
        \label{fig:objects}
    \end{figure}

    \begin{figure}
        \centering
        \includegraphics[width=\linewidth]{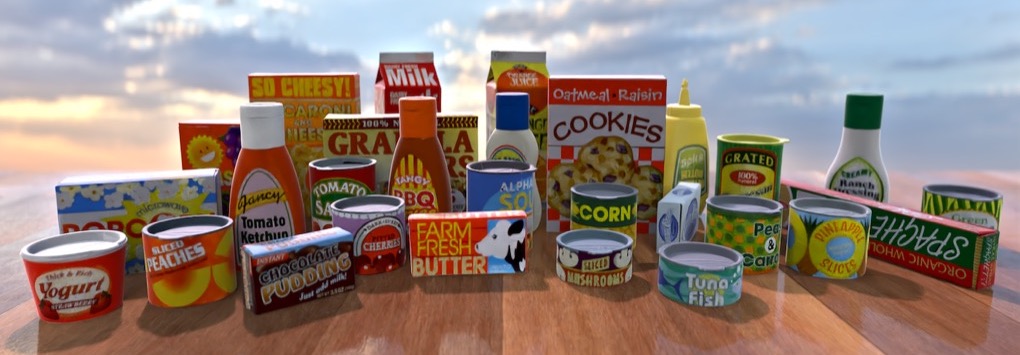}
        \caption{Synthetic rendering of 3D textured object models.}
        \label{fig:synobjects}
    \end{figure}
\else
\fi

Nevertheless, from a robotics point of view, another important limitation remains:  Existing pose estimators are trained %
for objects that are not available to most researchers.
This limitation is easily observed from the comprehensive list of datasets in the BOP Benchmark for 6D Object Pose Estimation~\cite{hodan2018eccv:bop}.\footnote{\url{https://bop.felk.cvut.cz/datasets/}}
Because there is no easy way to acquire most of the physical objects, robotics researchers are often unable to conduct real robotic experiments leveraging pose networks already trained using these datasets.

Of these existing datasets, arguably the most accessible are YCB~\cite{calli2015ram:ycb} and RU-APC~\cite{rennie2016ral:rutgersapc}.
However, over time it is becoming increasingly difficult to find matching objects from these datasets in stores.
For example, even though we live near the researchers who assembled the YCB dataset, we have been unsuccessful in locating a sugar box (YCB item 004) or gelatin box (YCB item 009) whose appearance matches that of the original.
One of the reasons for this failure is the changing appearance of such products. 
In particular, food manufacturers frequently change the design on their products based on seasonal or promotional considerations or updates to their branding.
As a result, networks trained on the YCB dataset exhibit degraded performance when applied to actual objects acquired from a store.
A further problem is that some of the items are not of the appropriate size, shape, and weight for common robotic grippers.

To address these problems, we release a dataset and benchmark for 6-DoF pose estimation for robotic manipulation research.
The dataset is designed to be:
\begin{itemize}
	\item \textbf{Accessible.}  The physical objects should ideally be accessible to anyone in the world.  As a result, we selected toy grocery objects that can be purchased online.  We also scanned these objects and share the resulting 3D textured models to aid in generating training images.
	\item \textbf{Challenging.}  We captured real images of these objects with the following characteristics:  1) significant occlusions and clutter, 2) varying number of instances of each object, 3) a wide variety of poses, and 4) extreme and challenging lighting conditions.
	\item \textbf{Accurate.}  We carefully labeled the images with ground truth poses, and errors in the labeled poses have been quantitatively assessed to be on the order of a few millimeters in world coordinates.
\end{itemize}
The dataset of test and validation images, with annotations for the latter, is available for download.
Additionally, metrics against the test set annotations are computed by the BOP Benchmark evaluation server. %
We include initial results 
using both DOPE~\cite{tremblay2018corl:dope} and CosyPose~\cite{labbe2020}, whose
pre-trained network weights are released as an off-the-shelf system for use in robotics labs as well as a baseline method for further research in this area. %
We also provide 3D object meshes for generating synthetic data for training.
We call our dataset HOPE, for Household Objects for Pose Estimation.

\section{Method}

In this section we describe the approach behind the dataset and benchmark.

\subsection{Set of Objects}

Because our goal is to support robotic manipulation research, it is imperative that the set of objects be:  1) somewhat realistic, 2) of the proper size and shape for grasping by a variety of robotic end effectors, and 3) accessible to researchers throughout the world.
An emerging area in robotic manipulation is that of household robots for automating daily chores in a laundry room or kitchen.
Such applications are crucial for facilitating aging-in-place, assisted living, and similar healthcare-related challenges.
While we initially intended to scan real items from a grocery store, we quickly realized that such an approach comes with several fundamental limitations:  1) such objects are not widely available to researchers in other parts of the world, 2) the reflective metallic properties of many of these objects make for difficult scanning and rendering, and 3) the surface textures of real-world objects frequently change due to seasonal marketing campaigns.
Again, it is not enough for us to provide a dataset for \emph{perception researchers} to design and test their algorithms---we also want to support \emph{robotics researchers} who need to leverage the latest perception research in their own labs.
As a result, the trained networks from the former should be immediately usable by the latter.

To this end, we chose a set of 28 toy grocery objects, depicted in Fig.~\ref{fig:objects}. These toys are widely available online and can be purchased for less than 60 USD total.
Because they are not real grocery items, there are no issues with perishability or transporting of the contents.
Moreover, these objects are of the proper size and shape for typical robotic grippers, with all objects having at least one dimension between 2.4 and 7.2~cm.

\ifsplashfig
\else
\begin{figure}
    \centering
    \includegraphics[width=\linewidth]{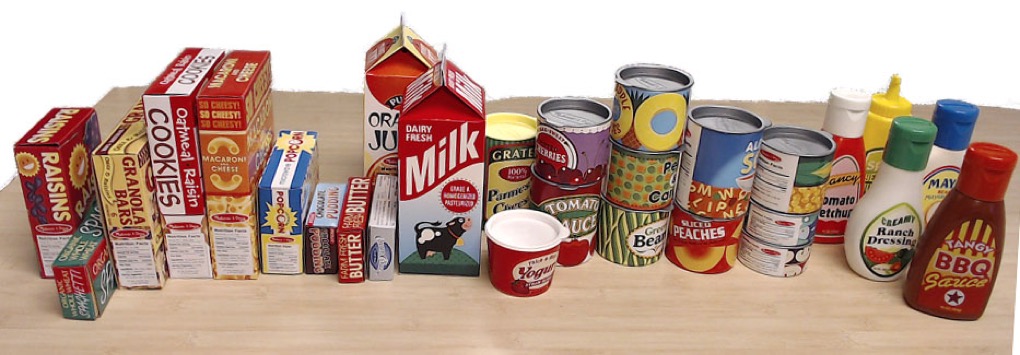}
    \caption{Set of 28 toy grocery objects used in the dataset.}
    \label{fig:objects}
\end{figure}

\begin{figure}
    \centering
    \includegraphics[width=\linewidth]{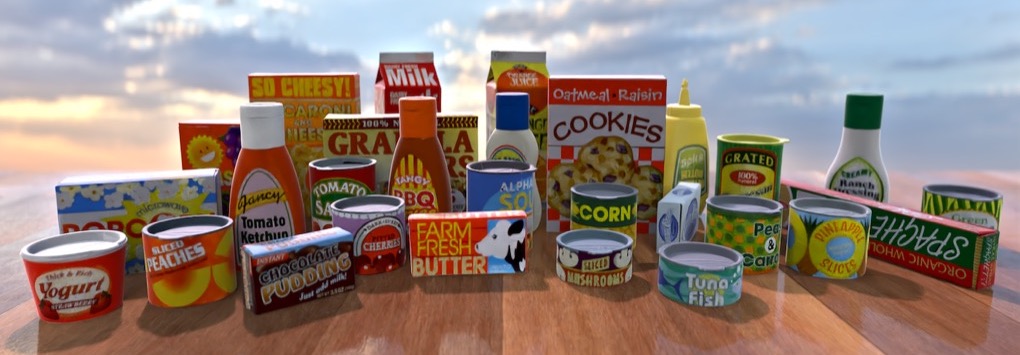}
    \caption{Synthetic rendering of 3D textured object models.}
    \label{fig:synobjects}
\end{figure}
\fi

\subsection{3D Textured Object Models}

Each object was scanned with a low-cost EinScan-SE desktop 3D scanner to generate textured 3D object meshes.
The scanning process introduced another limitation on both the size and materials of the objects, due to the inability of the scanner to handle objects larger than about 20~cm on any side.
The software tends to produce fragmented texture maps, so we used Maya to further refine the textures. %
Fig.~\ref{fig:synobjects} shows a synthetic rendering of the 3D textured object models,
rendered using NViSII, a Python-interfaced ray tracing library~\cite{Morrical2021iclrw:nvisii}.
Textured object models are useful for generating training data, as in DOPE~\cite{tremblay2018corl:dope} or BlenderProc~\cite{denninger2019blenderproc}.

\subsection{Capturing Real Images}

\begin{figure*}
    \centering
        \begin{subfigure}[b]{0.195\linewidth}
            \centering\includegraphics[width=\linewidth]{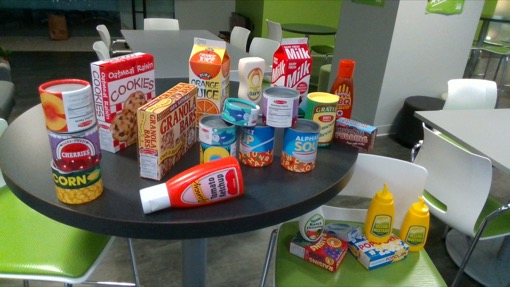}
            \vspace{-1.8em}
            \caption*{\scriptsize\sffamily break room}
        \end{subfigure}
        \begin{subfigure}[b]{0.195\linewidth}
            \centering\includegraphics[width=\linewidth]{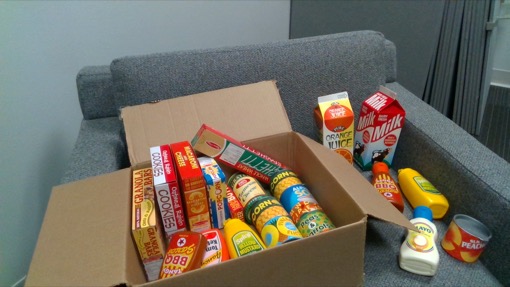}
            \vspace{-1.8em}
            \caption*{\scriptsize\sffamily chair}
        \end{subfigure}
        \begin{subfigure}[b]{0.195\linewidth}
            \centering\includegraphics[width=\linewidth]{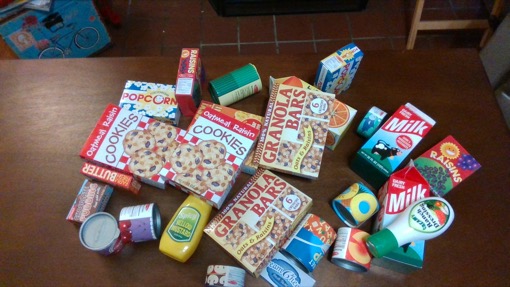}
            \vspace{-1.8em}
            \caption*{\scriptsize\sffamily coffee table}
        \end{subfigure}
        \begin{subfigure}[b]{0.195\linewidth}
            \centering\includegraphics[width=\linewidth]{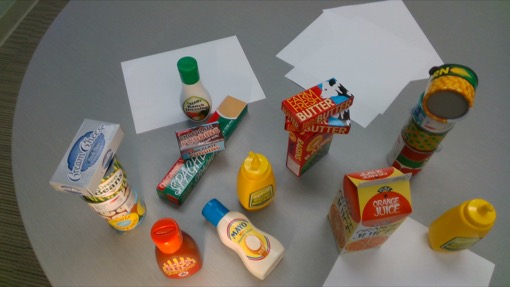}
            \vspace{-1.8em}
            \caption*{\scriptsize\sffamily conference table}
        \end{subfigure}
        \begin{subfigure}[b]{0.195\linewidth}
            \centering\includegraphics[width=\linewidth]{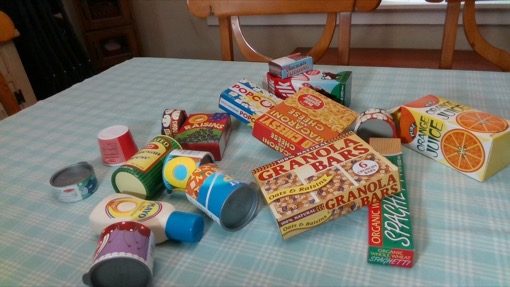}
            \vspace{-1.8em}
            \caption*{\scriptsize\sffamily dining table}
        \end{subfigure}
        \\\vspace{0.22em}
        \begin{subfigure}[b]{0.195\linewidth}
            \centering\includegraphics[width=\linewidth]{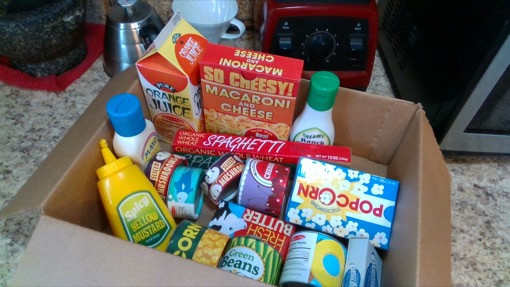}
            \vspace{-1.8em}
            \caption*{\scriptsize\sffamily kitchen}
        \end{subfigure}
        \begin{subfigure}[b]{0.195\linewidth}
            \centering\includegraphics[width=\linewidth]{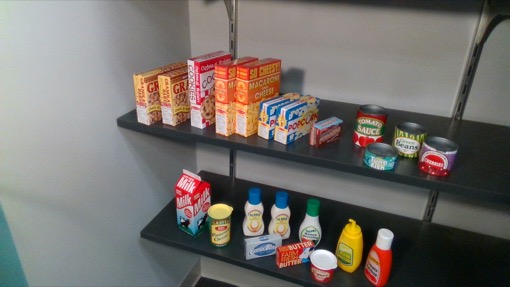}
            \vspace{-1.8em}
            \caption*{\scriptsize\sffamily pantry}
        \end{subfigure}
        \begin{subfigure}[b]{0.195\linewidth}
            \centering\includegraphics[width=\linewidth]{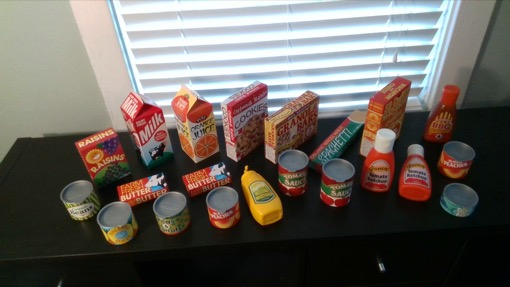}
            \vspace{-1.8em}
            \caption*{\scriptsize\sffamily window 1}
        \end{subfigure}
        \begin{subfigure}[b]{0.195\linewidth}
            \centering\includegraphics[width=\linewidth]{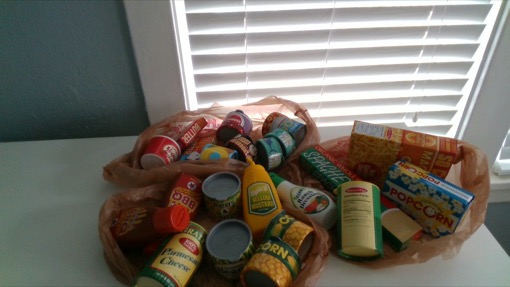}
            \vspace{-1.8em}
            \caption*{\scriptsize\sffamily window 2}
        \end{subfigure}
        \begin{subfigure}[b]{0.195\linewidth}
            \centering\includegraphics[width=\linewidth]{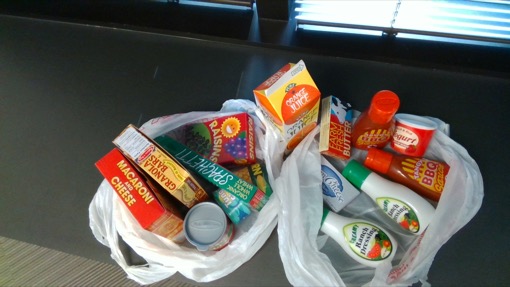}
            \vspace{-1.8em}
            \caption*{\scriptsize\sffamily window 3}
        \end{subfigure}
    \caption{Sample images from the ten environments. ``Chair'' and ``window 1'' are included in the validation set.}
    \label{fig:environments}
\end{figure*}

\begin{figure*}
    \centering
        \begin{subfigure}[b]{0.195\linewidth}
            \centering\includegraphics[width=\linewidth]{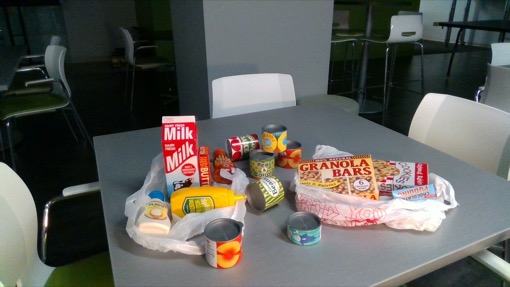}
            \vspace{-1.8em}
            \caption*{\scriptsize\sffamily bagged}
        \end{subfigure}
        \begin{subfigure}[b]{0.195\linewidth}
            \centering\includegraphics[width=\linewidth]{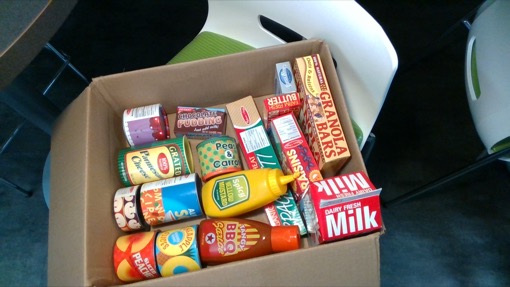}
            \vspace{-1.8em}
            \caption*{\scriptsize\sffamily boxed}
        \end{subfigure}
        \begin{subfigure}[b]{0.195\linewidth}
            \centering\includegraphics[width=\linewidth]{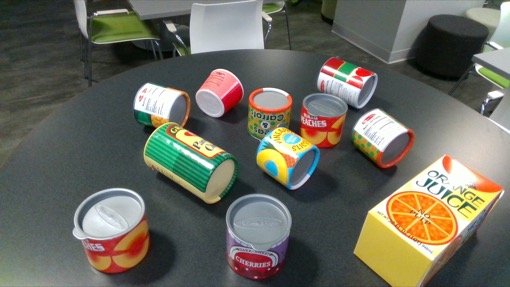}
            \vspace{-1.8em}
            \caption*{\scriptsize\sffamily dumped}
        \end{subfigure}
        \begin{subfigure}[b]{0.195\linewidth}
            \centering\includegraphics[width=\linewidth]{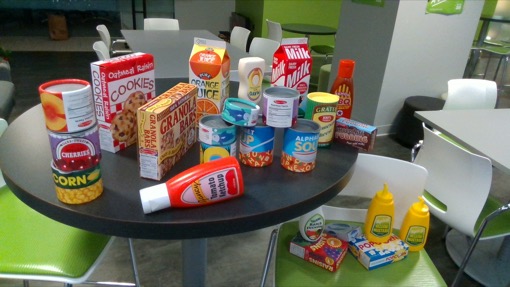}
            \vspace{-1.8em}
            \caption*{\scriptsize\sffamily stacked}
        \end{subfigure}
        \begin{subfigure}[b]{0.195\linewidth}
            \centering\includegraphics[width=\linewidth]{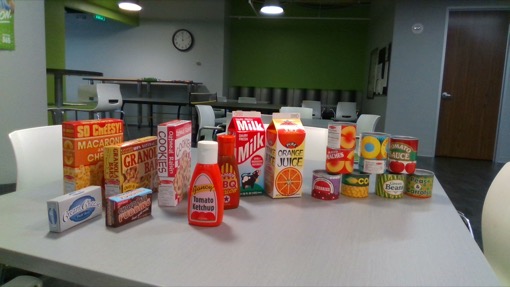}
            \vspace{-1.8em}
            \caption*{\scriptsize\sffamily tidy}
        \end{subfigure}
    \caption{Five object arrangements, shown for the ``break room'' environment.}
    \label{fig:arrangements}
\end{figure*}

To acquire images for the dataset, we placed the real objects in ten different environments, with five object arrangements / camera poses per environment.
The environments are shown in Fig.~\ref{fig:environments}, and the object arrangements for one environment are depicted in Fig.~\ref{fig:arrangements}.
These 50 different scenes exhibit a wide variety of backgrounds, clutter, poses, and lighting.
With some arrangements, objects are placed inside other containers (\emph{e.g.,} boxes, bags, or drawers) to provide additional clutter and partial occlusion.
Both RGB and depth images were collected for each scene using an Intel RealSense D415 RGBD camera at full HD resolution ($1920 \times 1080$).
Images were captured from distances of 0.5 to 1.0~m, which are typical of robotic grasping.

\begin{figure}
    \centering
		\begin{tabular}{cc}
		\hspace{-0.05in}\includegraphics[width=0.48\linewidth]{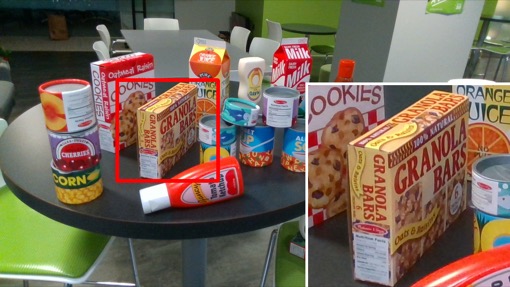} &
		\hspace{-0.1in}\includegraphics[width=0.48\linewidth]{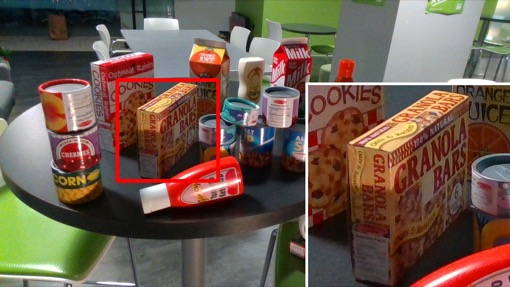} \\
		\hspace{-0.05in}\includegraphics[width=0.48\linewidth]{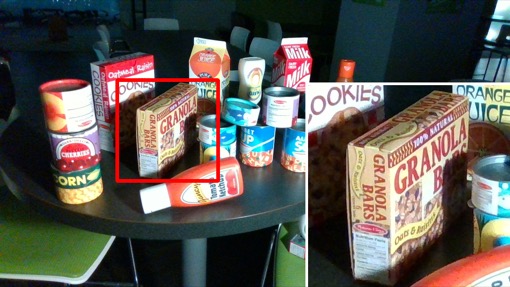} &
		\hspace{-0.1in}\includegraphics[width=0.48\linewidth]{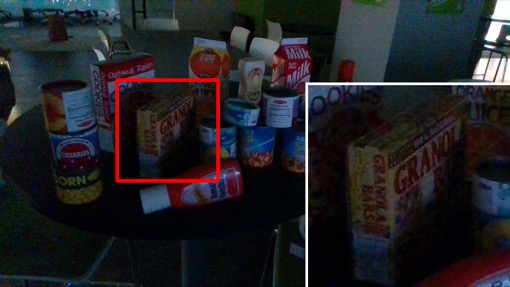}
		\end{tabular}
    \caption{Four lighting variations of the same scene. Insets show zoomed crops.}
    \label{fig:lighting}
\end{figure}

Once the objects were arranged in an environment and the camera was set, we acquired multiple images with various lighting conditions by turning on/off the lights, opening window blinds, and so forth.
Since the scene was static, the annotations (described below) do not change and thus required no additional work.
In this manner we were able to collect a large variety of lighting conditions, see Fig.~\ref{fig:lighting}.

In total, the dataset contains 50 unique scenes, 238 images (an average of 4.8 lighting variations per scene), and 914 object poses.
Images from two environments (``chair'') and (``window 1'') are made available with ground truth for a validation set, while the remaining scenes are reserved as the test set.
\ifarxiv
Fig.~\ref{fig:distribution_pose} summarizes the object sizes, as well as the  variety of object pose angles, estimated visibility,
and distance to the camera across all pose instances in the entire dataset.
\fi
Minimum and maximum object dimensions are in the ranges 2.4--7.2~cm and 6.8--25.0~cm, respectively.
Average object visibility is 83\%, with only 39\% of objects in unoccluded poses ($>$95\% visible).
Objects are set in a wide variety of orientations, and object distances range from 39.0--141.0~cm with an average distance of 73.8~cm.

\subsection{Annotating Images with Ground Truth} %

We annotated the dataset with ground truth object poses by manually identifying point correspondences between images and 3D textured object models.
In the P$n$P version of our annotation tool, the annotator selects corresponding points in the 2D texture map of the object model and the image in which the object appears, after which the 3D model is aligned to the image by P$n$P with RANSAC.  Alternatively, in the RGBD version, correspondences are made between the 3D textured model and the 2.5D RGBD depth map, with alignment made by Procrustes.
The RGBD annotation tool is typically faster to operate but may suffer from noise or bias in depth measurements, while P$n$P is more error-prone along the projection ray from the camera to the object. Most annotations were performed with the RGBD tool, resorting to P$n$P only when the depth-based annotation was not satisfactory.
All annotations were automatically refined by SimTrack~\cite{pauwels2015iros:simtrack}.
Then, if necessary, the alignment was refined manually %
until the reprojected model was visually aligned with both the image and the depth map.

\ifarxiv
\begin{figure}
    \centering
        \begin{subfigure}[b]{0.49\linewidth}
            \centering\includegraphics[trim=2px 2px 2px 2px, clip, width=\linewidth]{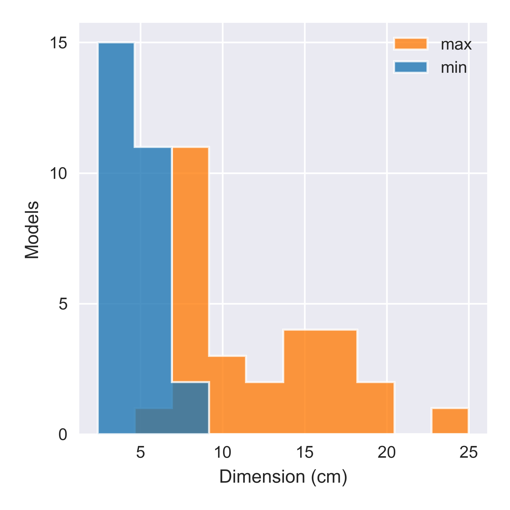}
            \caption*{\scriptsize\sffamily (a) Object max/min dim.}
        \end{subfigure}
        \begin{subfigure}[b]{0.49\linewidth}
            \centering\includegraphics[trim=2px 2px 2px 2px, clip, width=\linewidth]{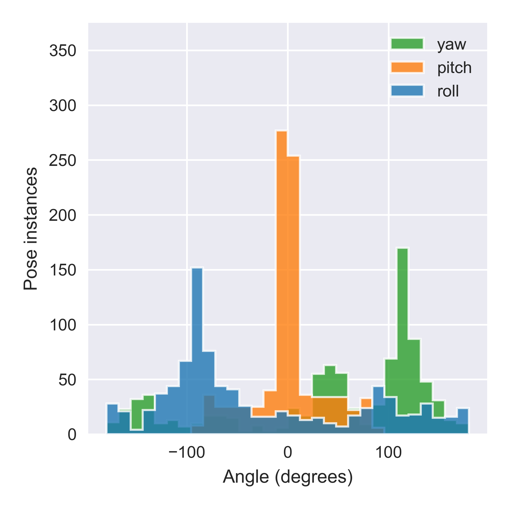}
            \caption*{\scriptsize\sffamily (b) Pose angle}
        \end{subfigure}
        \\\vspace{0.3em}
        \begin{subfigure}[b]{0.49\linewidth}
            \centering\includegraphics[trim=2px 2px 2px 2px, clip, width=\linewidth]{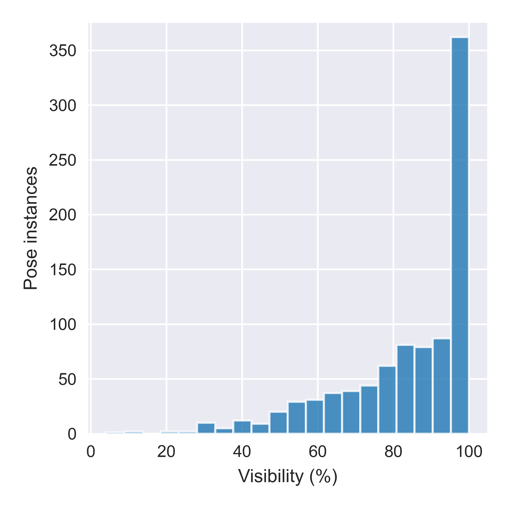}
            \caption*{\scriptsize\sffamily (c) Pose visibility}
        \end{subfigure}
        \begin{subfigure}[b]{0.49\linewidth}
            \centering\includegraphics[trim=2px 2px 2px 2px, clip, width=\linewidth]{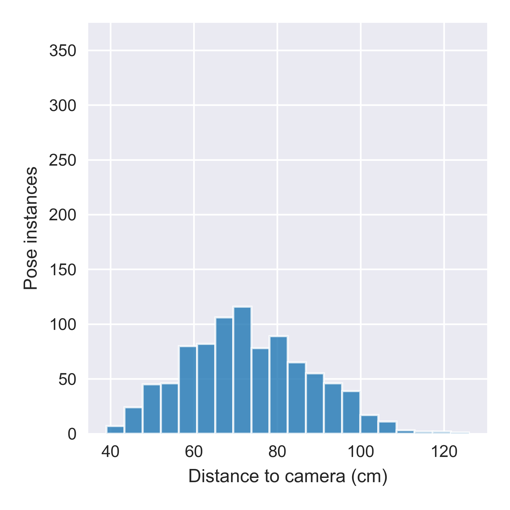}
            \caption*{\scriptsize\sffamily (d) Pose distance}
        \end{subfigure}
    \caption{Distribution of object sizes, and their visibility and poses in the dataset.}
    \label{fig:distribution_pose}
\end{figure}
\fi

\subsection{Depth Calibration}

Following Hoda{\v{n}} et al.~\cite{hodan2017wacv:tless}, we checked the depth calibration of the RealSense camera and found a small systematic error. We captured over 50 images of a checkerboard at distances between 0.5 and 2~m using the RealSense left-infrared camera.
Checkerboard corners were automatically detected, then unprojected to 3D using P$n$P and the camera intrinsics.
The resulting coordinates were compared with the corresponding 3D coordinates obtained from the 2.5D depth map.
Fitting a scale factor of 0.9804 reduced the mean absolute difference between measurements from 19.3 to 7.6~mm.
The depth images included in the dataset were scaled by this factor before registering to RGB.

\subsection{Symmetry-Aware Metrics}

Perhaps the most common and easily interpretable metric for evaluating pose estimates is the average distance (ADD) metric~\cite{hinterstoisser2012accv:linemod}, which computes the mean pairwise distance between corresponding vertices in the 3D object model under the ground truth ($\bar\bP$) and predicted ($\hat\bP$) poses:
\begin{align}
	e_{\text{ADD}} &= \underset{\bx \in O}{\mathrm{mean}} \hspace{1mm} \| \bar\bP\bx - \hat\bP\bx\|,
	\label{eq:eadd}
\end{align}
where $\bx \in O$ are the positions of model vertices in the object coordinate frame, and $\|\cdot\|$ is the $\ell_2$-norm.

Because ADD makes no allowances for object symmetries, it provides only limited insight into the \emph{graspability} of detected objects based on the predicted pose, as translation errors are penalized similarly to rotation errors.
Moreover, in the case of symmetric objects, ADD penalizes predictions even when the true pose cannot be determined from the input image.
To overcome these limitations, some researchers adopt a mean closest point distance (ADD-S)~\cite{xiang2018rss:posecnn}:
\begin{align}
	e_{\text{ADD-S}} &= \underset{\bx_1 \in O}{\mathrm{mean}} \hspace{1mm} \min_{\bx_2 \in O} \| \bar\bP\bx_1 - \hat\bP\bx_2\|,
\end{align}
where each vertex in the ground truth model is paired with the \emph{nearest} vertex in the prediction model, irrespective of consistency with other assignments. As shown later in this section, this metric can significantly underestimate pose error due to unrealistic pairings.

Another way to handle symmetries is to directly model the set $\bS_O$ of symmetry transformations corresponding to the object.
Starting from ADD in Eq.~\eqref{eq:eadd}, this leads to the Mean Symmetry-Aware Surface Distance (MeanSSD):
\begin{align}
	e_{\text{MeanSSD}} &= \min_{\bS \in \bS_O} \underset{\bx \in O}{\mathrm{mean}} \hspace{1mm} \| \bar\bP\bS \bx - \hat\bP \bx\|.
	\label{eq:meanssd}
\end{align}
MeanSSD requires calculating or manually identifying all valid (discrete and/or continuous) symmetries for a given object, a potentially burdensome task for a large and varied set of objects.
If the mean operator in Eq.~\eqref{eq:meanssd} is replaced with max, MeanSSD becomes the Maximum Symmetry-Aware Surface Distance (MSSD) proposed by Hoda{\v{n}} et al.~\cite{hodan2020eccv:bop}:
\begin{align}
	e_{\text{MSSD}} &= \min_{\bS \in \bS_O} \max_{\bx \in O} \hspace{1mm} \| \bar\bP\bS \bx - \hat\bP \bx\|.
	\label{eq:mssd}
\end{align}

As a compromise between the inaccurate ADD-S metric and the labor-intensive MeanSSD/MSSD metrics, we propose a variant of the average distance metric in which the vertex correspondences between ground truth and predicted models are made by solving a linear sum assignment problem. The most famous solution to this assignment problem is the Hungarian algorithm~\cite{crouse2016taes:assignment}, hence we denote this metric ADD-H.
The assignment $f_A: \bx_2 \mapsto \bx_1$ produces a bijective mapping
from the vertices of the model in the predicted pose to those in the ground truth pose.  If we let $A=\{(f_A(\bx_2), \bx_2)\}$ be the set of such correspondences, then the set that minimizes the sum of distances between paired vertices is calculated as:
\begin{align}
	 & \tilde A = \arg\min_{A} \sum_{(\bx_1,\bx_2) \in A} \hspace{1mm} \| \bar\bP\bx_1 - \hat\bP\bx_2\|.
     \label{eq:addhassign}
\end{align}
We compute Eq.~\eqref{eq:addhassign} using a modern variation of the Hungarian algorithm.\footnote{Implemented as \texttt{scipy.optimize.linear\_sum\_assignment}.  Note that versions of SciPy before 0.17.0 use a much slower algorithm.} The average distance is then computed between assigned pairs $(\bx_1,\bx_2) \in \tilde A$:
\begin{align}
	e_{\text{ADD-H}} &= \underset{(\bx_1,\bx_2) \in \tilde A}{\mathrm{mean}} \hspace{1mm} \| \bar\bP\bx_1 - \hat\bP\bx_2\|.
    \label{eq:addh}
\end{align}

\begin{figure}
    \centering
    \centering\includegraphics[trim=0px 0px 0px 0px, clip, width=\linewidth]{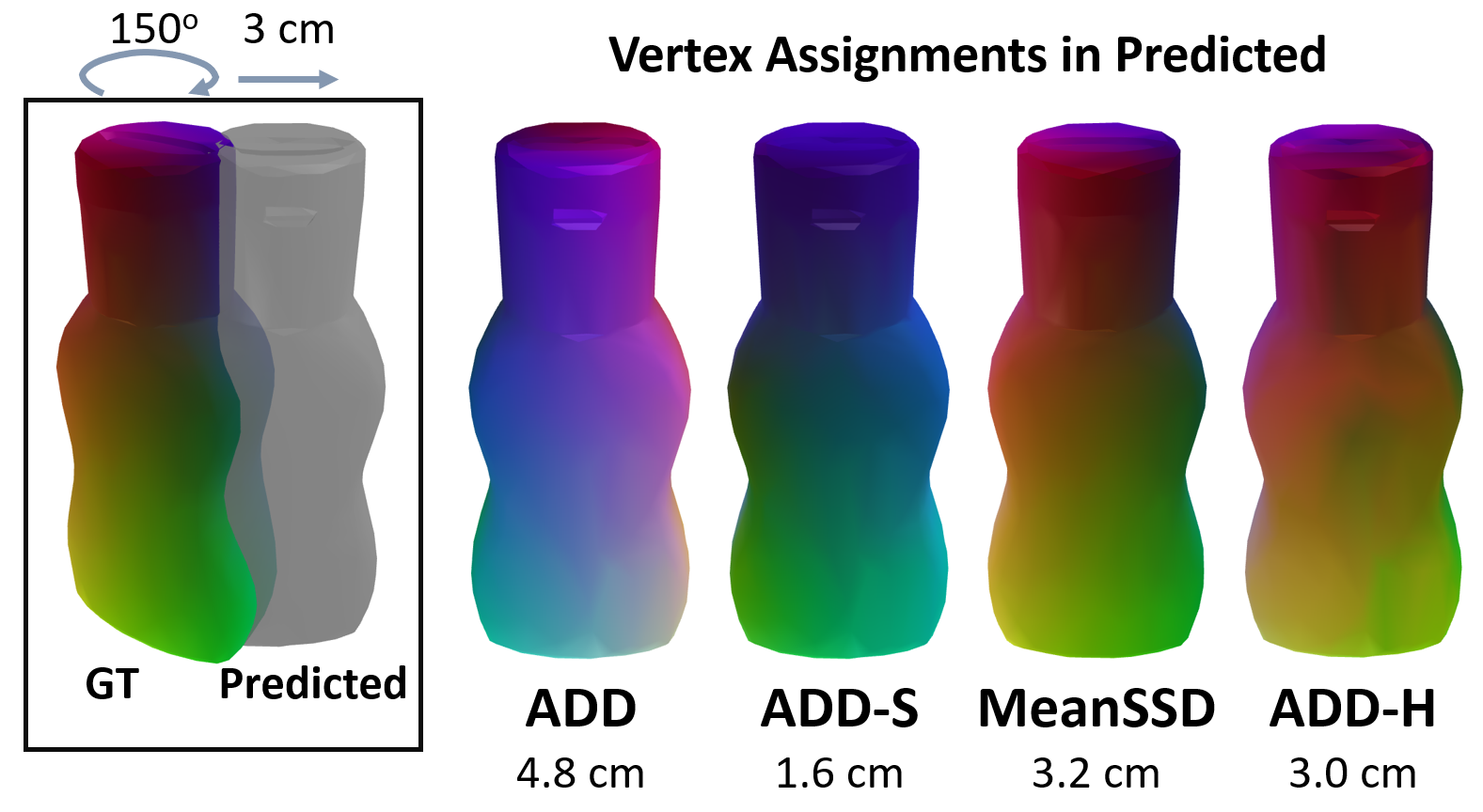}
    \caption{Pairwise vertex assignments (shown by vertex color) between ground truth (GT) and predicted model poses using four different metrics.}
    \label{fig:metricviz}
\end{figure}

\begin{figure}
    \centering
    \begin{subfigure}[b]{0.49\linewidth}
        \centering\includegraphics[trim=2px 3px 2px 2px, clip, width=\linewidth]{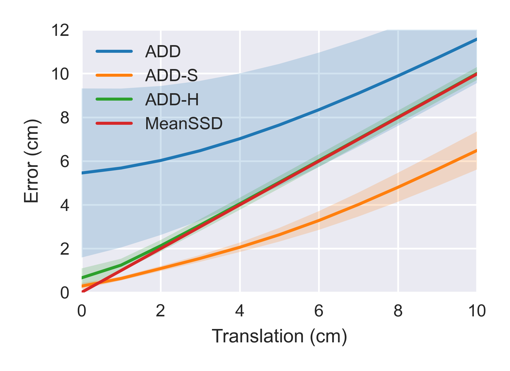}
        \caption*{(a) Symmetry-preserving}
    \end{subfigure}
    \begin{subfigure}[b]{0.49\linewidth}
        \centering\includegraphics[trim=2px 3px 2px 2px, clip, width=\linewidth]{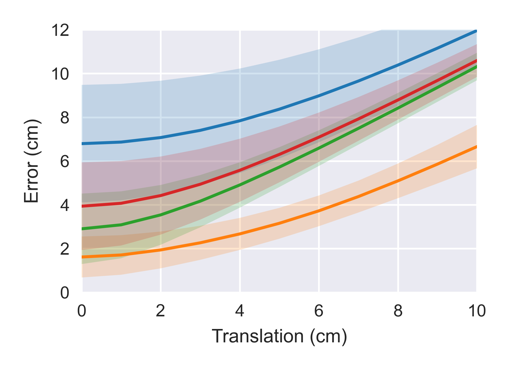}
        \caption*{(b) Non-symmetry-preserving}
    \end{subfigure}
    \caption{Error metrics ($\pm$1 standard deviation) between object poses with random rotations and increasing translation.  ADD-H closely approximates MeanSSD.}
    \label{fig:metricsimulation}
\end{figure}

Fig.~\ref{fig:metricviz} visualizes the vertex correspondences used by each metric when computing the error between the ground truth and predicted pose for a bottle. The inset rendering on the left depicts the scene, where the predicted pose (in translucent gray) is offset from the ground truth pose (with colored vertices) by a 150{\degree} rotation about the vertical axis and a horizontal translation of about half the object width (3\,cm). %

The right side of the figure depicts the vertex assignments for each metric: vertices in each model ($\bx_2 \in O$) are colored according to the paired vertices in the ground truth model ($\bx_1 \in O$).
Since ADD uses a fixed, bijective 
assignment based on the original vertex ordering, %
the error is large. %
In contrast, ADD-S assigns the nearest ground truth vertex to each target, resulting in a non-injective mapping,
in which only points from the right side of the ground truth are used, as reflected in the green and purple color across the entire target mesh.
MeanSSD uses the same vertex assignment as ADD but explicitly selects the symmetry-preserving rotation that minimizes error, %
resulting in a rotational alignment that is more representative of the grasp-relevant geometry of the prediction.
Finally, ADD-H optimizes a bijective mapping between poses to minimize error---achieving a vertex assignment that reflects the pose symmetry, similar to MeanSSD.

To further compare the metrics, Fig.~\ref{fig:metricsimulation} shows the error, averaged over 5 trials for each of the 28 HOPE objects, as each object is translated from its initial position in increments of 1~cm.
In the left plot, each object is first randomly rotated by one of the symmetry-preserving transformations, whereas in the right plot,
each object is first randomly rotated by an arbitrary transformation.
ADD-H closely matches the MeanSSD error without any explicit enumeration of object symmetries, especially when the rotation is symmetry-preserving.
In contrast, ADD and ADD-S significantly over- and under-estimate the error, respectively.

Although ADD-H is computationally more expensive than related methods, 
by using an efficient algorithm to solve the assignment problem it can be applied to meshes with several hundred vertices.
In our experiments, we use 500 vertices, which yields consistent results with sufficiently fast computation.

\section{Experiments}

We first present a validation experiment that examines the accuracy of our annotations.
Then we provide a baseline experiment where we train several pose detectors and report accuracy metrics for the dataset.

\subsection{Annotation Validation Experiment}

\begin{figure}
    \centering
    \begin{subfigure}{0.48\linewidth}
        \centering\includegraphics[trim=20px 20px 20px 20px, clip, width=\linewidth]{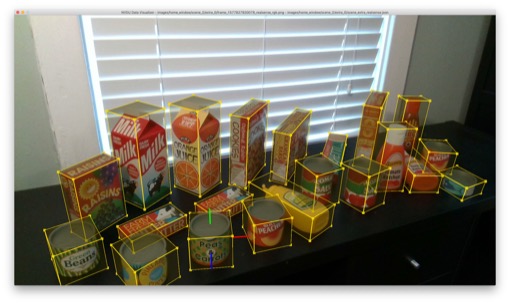}
    \end{subfigure}
    \begin{subfigure}{0.48\linewidth}
        \centering\includegraphics[trim=20px 20px 20px 20px, clip, width=\linewidth]{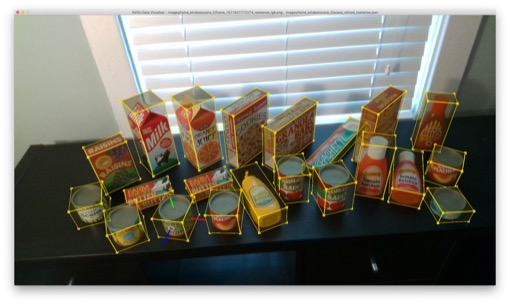}
    \end{subfigure}
    \caption{Two views for computing accuracy via consistency.}
    \label{fig:twoviews}
\end{figure}

\begin{figure}
    \centering
    \includegraphics[trim=2.5px 2.5px 2.5px 3px, clip, width=0.85\linewidth]{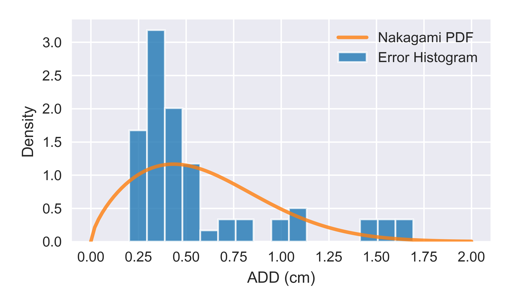}
    \caption{Histogram of ADD errors in annotation validation experiment with best fit Nakagami distribution ($m\!=\!0.831, \Omega\!=\!0.830$, mean$\,=\!0.597$, mode$\,=\!0.435$).\protect\footnotemark}
    \label{fig:distribution_add}
\end{figure}

\begin{figure*}
    \setlength{\tabcolsep}{0pt}
    \begin{tabular}{cccc}
        \multicolumn{4}{c}{\scriptsize\sffamily \textbf{Validation Set}} \\
        \vspace{-0.25em}
        \begin{subfigure}[b]{0.24\linewidth}
            \centering\includegraphics[trim=4px 4px 4px 4px, clip, width=\linewidth]{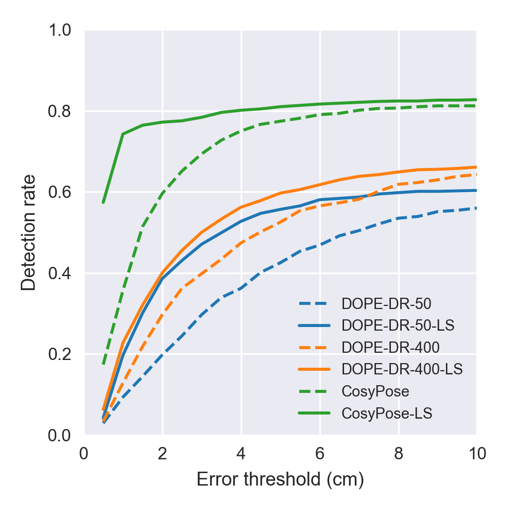}
        \end{subfigure} &
        \begin{subfigure}[b]{0.24\linewidth}
            \centering\includegraphics[trim=4px 4px 4px 4px, clip, width=\linewidth]{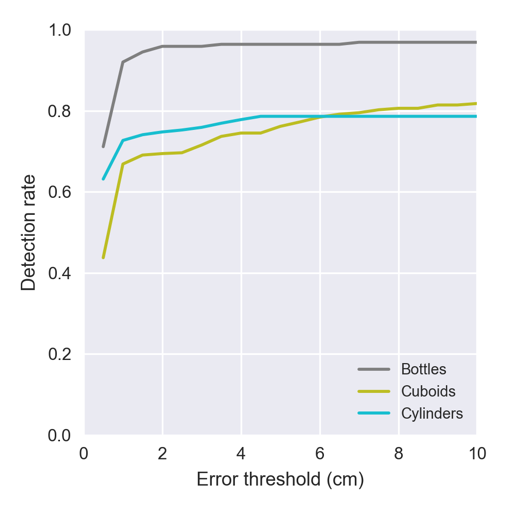}
        \end{subfigure} &
        \begin{subfigure}[b]{0.24\linewidth}
            \centering\includegraphics[trim=4px 4px 4px 4px, clip, width=\linewidth]{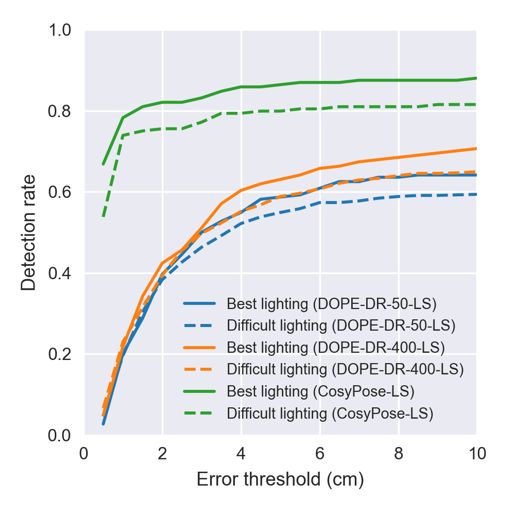}
        \end{subfigure} &
        \begin{subfigure}[b]{0.24\linewidth}
            \centering\includegraphics[trim=4px 4px 4px 4px, clip, width=\linewidth]{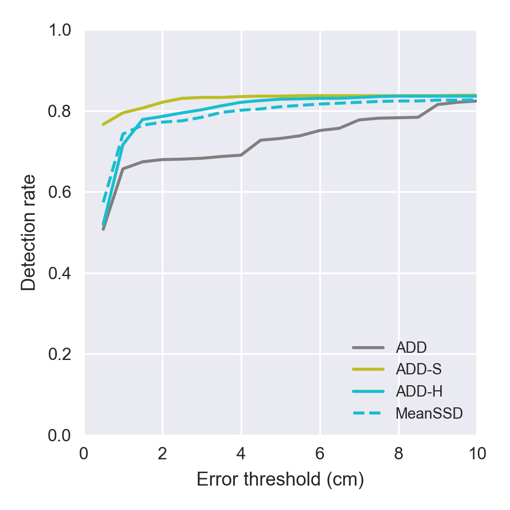}
        \end{subfigure} \\
        \multicolumn{4}{c}{\scriptsize\sffamily \textbf{Test Set}} \\
        \begin{subfigure}[b]{0.24\linewidth}
            \centering\includegraphics[trim=4px 4px 4px 4px, clip, width=\linewidth]{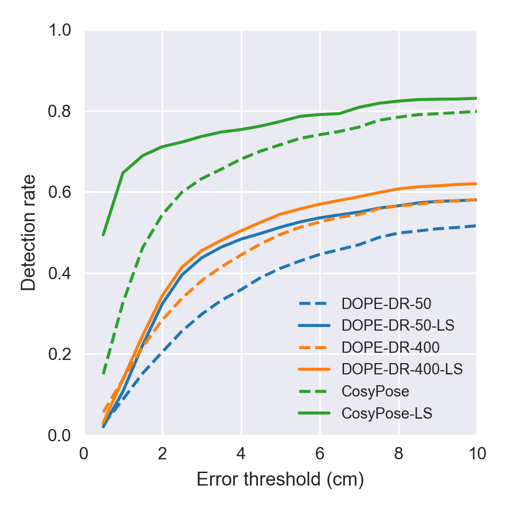}
        \end{subfigure} &
        \begin{subfigure}[b]{0.24\linewidth}
            \centering\includegraphics[trim=4px 4px 4px 4px, clip, width=\linewidth]{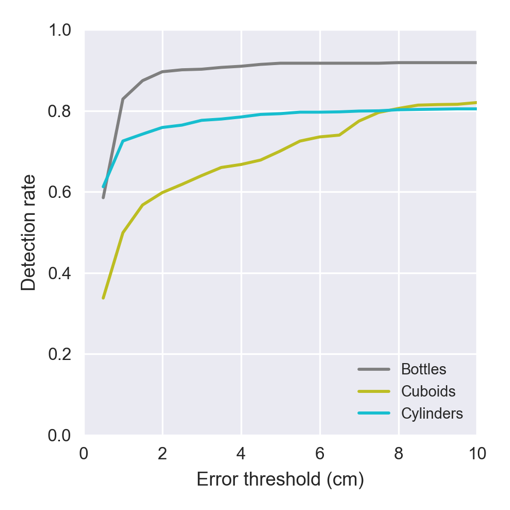}
        \end{subfigure} &
        \begin{subfigure}[b]{0.24\linewidth}
            \centering\includegraphics[trim=4px 4px 4px 4px, clip, width=\linewidth]{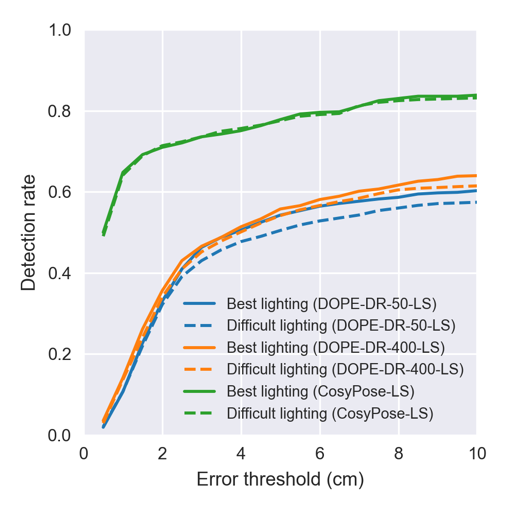}
        \end{subfigure} &
        \begin{subfigure}[b]{0.24\linewidth}
            \centering\includegraphics[trim=4px 4px 4px 4px, clip, width=\linewidth]{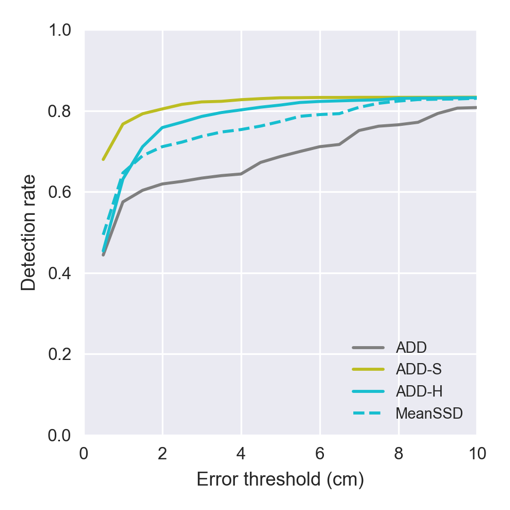}
        \end{subfigure} \\

        \vspace{-0.25em}
        \footnotesize{(a) Methods} &
        \footnotesize{(b) Object categories} &
        \footnotesize{(c) Lighting} &
        \footnotesize{(d) Metrics} \\

        \scriptsize{(\defaultmetric{})} &
        \scriptsize{(CosyPose-LS, \defaultmetric{})} &
        \scriptsize{(\defaultmetric{})} &
        \scriptsize{(CosyPose-LS)} \\
    \end{tabular}
    \caption{Detection rate versus maximum detection threshold, computed on the HOPE validation set (top row) and test set (bottom row).}
    \label{fig:meanssdresults}
\end{figure*}

To estimate the error in our ground truth pose annotations, we captured orthogonal views ($\sim\!90$ degrees between camera axes) of three static scenes.
Objects were annotated independently in both views, as shown in Fig.~\ref{fig:twoviews}.
Using a robust Procrustes alignment of all but one annotated object pose, we estimated the extrinsics between the two camera views.
The pose of the held-out object was projected from the first view to the second, and ADD error was computed between the transferred pose and the annotation made directly in the second view.
By repeating the process while holding out each object in turn, we estimated the annotation error for all objects in the scene.
Fig.~\ref{fig:distribution_add} shows a histogram of ADD for the 64 object instances contained in the three scenes. The mean and median ADD are 5.7 and 4.3~mm, respectively.
We conclude, therefore, that ground truth is accurate to several millimeters.

\footnotetext{
If the vertices in two random poses have Gaussian errors, then the sum of their squared differences is given by a gamma distribution.  
If $z$ follows a gamma distribution with parameters $\alpha$ and $\beta$, then $\sqrt{z}$ follows a Nakagami distribution with parameters $m=\alpha$ and $\Omega=\alpha\beta$.
Thus the mean Euclidean distance in Eq.~\eqref{eq:eadd} is expected to follow a Nakagami distribution.}

\subsection{Pose Prediction Baselines}

We trained baseline pose prediction models for each object class using the Deep Object Pose Estimation (DOPE)~\cite{tremblay2018corl:dope} and CosyPose~\cite{labbe2020} methods for detecting and predicting object poses in RGB images. For DOPE, models were trained from synthetic images using domain randomization~\cite{tobin2017iros:dr}. Unlike the original DOPE paper, we did not train using any photorealistic rendered images, which no doubt degrades performance.  Accordingly, we denote the baseline as DOPE-DR. Two architecture variants are considered with $50 \times 50$ and $400 \times 400$ output maps, denoted DOPE-DR-50 and DOPE-DR-400, respectively. For CosyPose, we trained using photorealistic images rendered by the BlenderProc tool~\cite{denninger2019blenderproc}.

As both DOPE and CosyPose are RGB-only methods, both can suffer from errors along the projection ray from the camera to the object. We consider a simple method for refinement using the RGB-D point clouds: given a predicted pose, visible model vertices are aligned with the depth map %
by adjusting the predicted translation by a scalar factor determined by a line search.
We denote this method with ``-LS''.
CosyPose has a distinct advantage in that the object mask from the Mask R-CNN detector can be used to filter both the object mesh and RGB-D point cloud, thereby limiting the effect of occluding objects.

\subsection{BOP Challenge}

\begin{table}
\centering
\caption{Average recall under three BOP Challenge metrics for the HOPE test set.}\label{tab:bopresults}
\begin{tabular}{llccc}
    \toprule
    \tH{Method} & \tH{Modality} & \txtsub{\tH{AR}}{\tH{MSSD}} & \txtsub{\tH{AR}}{\tH{VSD}} & \txtsub{\tH{AR}}{\tH{MSPD}} \\
    \midrule
    DOPE-DR-50     & RGB   & 0.232 & 0.215 & 0.457 \\
    DOPE-DR-50-LS  & RGB-D & 0.298 & 0.258 & 0.439 \\
    DOPE-DR-400    & RGB   & 0.297 & 0.277 & 0.498 \\
    DOPE-DR-400-LS & RGB-D & 0.317 & 0.285 & 0.470 \\
    CosyPose       & RGB   & 0.476 & 0.550 & \textbf{0.637} \\
    CosyPose-LS    & RGB-D & \textbf{0.594} & \textbf{0.691} & 0.629 \\
    \bottomrule
\end{tabular}
\end{table}

Table~\ref{tab:bopresults} shows the average recall (AR) under three metrics \emph{as computed by the BOP benchmark}~\cite{hodan2018eccv:bop}: Maximum Symmetry-Aware Surface Distance (MSSD), Visible Surface Discrepancy (VSD), and Maximum Symmetry-Aware Projection Distance (MSPD). 
MSSD, which measures 3D alignment of model vertices, is the most relevant to robotics applications.
VSD measures the discrepancy between 2D distance maps, rendered for GT and predicted poses, respectively. 
MSPD measures the discrepancy between 2D projections of model vertices, thereby capturing visual alignment with an eye toward augmented reality applications.

CosyPose-LS yields a nearly 2x improvement in \txtsub{AR}{MSSD} over DOPE-400-LS.
Furthermore, mistakes made along the projection ray to the object are shown to be a significant source of error, as evidenced by the dramatic improvement made by our simple depth refinement (-LS).
The refinement improves MSSD and VSD scores dramatically for all three methods, including a 25\% improvement in CosyPose.
Because MSPD is sensitive to subtle changes in visual alignment, it suffers slightly under depth refinement despite overall better 3D alignment.

Two aspects of the BOP metrics are worth noting: 1) as mentioned previously, the BOP definition of symmetries requires both geometric and (nearly) exact visual similarity, meaning that none of the HOPE objects are considered symmetric in the BOP challenge; and 2) the recall error threshold is defined \emph{relative} to object size, varying from 5\% to 50\% of the object diameter~\cite{hodan2020eccv:bop}.

\subsection{Detailed Experiments}

Because of our focus on graspability, we present additional results using \emph{absolute} detection thresholds, varying from 0.5 to 10~cm.
We also consider all of our objects to be geometrically symmetric, ignoring small details like tabs on can lids. %
For defining geometric symmetries, meshes were manually aligned to a consistent set of coordinate axes and grouped into three categories:
\emph{cylinders} (360{\degree} rotation around vertical axis in 1{\degree} increments, and 180{\degree} flip top to bottom), \emph{cuboids} (180{\degree} flip along any axis), and \emph{bottles} (180{\degree} flip front to back).

Fig.~\ref{fig:meanssdresults} presents the results of four experiments, depicting the detection rate of ground truth objects (recall) on the $y$-axis while varying the maximum error threshold for a positive detection on the $x$-axis. 
We show results on both validation and test sets.
To facilitate interpretability by those familiar with existing metrics, we use the \defaultmetric{} metric.\footnote{
For versions of Fig.~\ref{fig:meanssdresults}\,(a-c) and Table~\ref{tab:meanssdresults} using the ADD-H metric, see
\ifarxiv
Fig.~\ref{fig:addhresults} and Table~\ref{tab:addhresults}, respectively, in the Appendix.
\else
the Appendix of the \href{https://arxiv.org/abs/2203.05701}{arXiv version} of this paper.
\fi
}

In the analysis that follows, we specifically highlight performance at two particular error thresholds: 2~cm and 10~cm.
The smaller threshold (2~cm) approximately indicates errors sufficient for grasping, although this will of course vary by gripper.
The larger threshold (10~cm) indicates errors sufficient for a system that is able to gather additional views of a detected object for refined results, \emph{e.g.,} when the camera is in the robot hand---although this threshold also is an approximation.

Fig.~\ref{fig:meanssdresults}\,(a) compares CosyPose and DOPE using the \defaultmetric{} metric.
CosyPose-LS, trained with photorealistic images, incorporating a visual refinement step, and augmented with our simple depth refinement, predicts over 70\% of objects in the HOPE test set within 2~cm.
Considering DOPE, the figure shows the orthogonal benefits of a larger output map (DOPE-DR-400-*)---which improves the precision of keypoint prediction in the DOPE method---and line search depth refinement (DOPE-DR-*-LS). With line search and depth refinement, more than 30\% of objects in the HOPE test set are predicted within 2~cm.

Fig.~\ref{fig:meanssdresults}\,(b) shows the performance of CosyPose-LS using the \defaultmetric{} metric across three object categories. Cuboids present a particular challenge since they are less resistant to small rotational errors. In this dataset, cuboids also tend to be larger objects that are placed further from the camera and behind other objects in some scene arrangements.
Highest success is obtained with bottles, with almost 90\% predicted within 2~cm.

Fig.~\ref{fig:meanssdresults}\,(c) depicts the effect of lighting variations by comparing detection rate (again using \defaultmetric{}) between the most favorable image and most difficult image in each scene (determined subjectively). Particularly on the test set, which is much larger than the validation set, the difference in accuracy is small, indicating a promising level of robustness in these synthetically trained methods.

Finally, Fig.~\ref{fig:meanssdresults}\,(d) compares the various metrics using CosyPose-LS predictions, confirming previous observations that ADD-H is a reasonable compromise that matches the results of MeanSSD without requiring explicit enumeration of object symmetries. 
As mentioned earlier, ADD over-estimates, while ADD-S under-estimates, the error.

Finally, Table~\ref{tab:meanssdresults} presents detailed statistics of CosyPose-LS predictions for each object class and category, using the \defaultmetric{} metric, along with some statistics of the ground truth (GT) object poses.
The table includes the median of the \defaultmetric{} error among objects detected within a 10~cm threshold, as well as precision and recall at both 2~cm and 10~cm thresholds. 
At the most generous 10~cm detection threshold, 83\% of ground truth objects (in the test set) are detected by CosyPose-LS,
and at the tighter 2~cm threshold, 72\% are detected, meaning that grasping may be feasible.
CosyPose-LS could benefit from an improved object detection step, as nearly 20\% of objects are not detected.
False positives appear to be less of a problem, with $>$98\% precision at the 10~cm threshold. %
These results offer room for improvement, while showing that an existing method like CosyPose can be reasonably successful as an off-the-shelf predictor for these objects.

\begin{table*}
\centering
\caption{Results on the validation and test sets using CosyPose with the \defaultmetric{} metric. (The median error was computed among all ``true positive'' predictions, determined at a 10~cm error threshold.)
}
\label{tab:meanssdresults}
\resizebox{0.97\textwidth}{!}{%
\begingroup
\setlength{\tabcolsep}{5pt} %
\renewcommand{\arraystretch}{1.1} %
\rowcolors{6}{gray!10}{white!0}
\begin{tabular}{@{}rlcccccccccccccc} %

    \toprule
     & & & \multicolumn{3}{c}{\tH{Pose Statistics}} & \multicolumn{5}{c}{\tH{Validation Set}} & \multicolumn{5}{c}{\tH{Test Set}} \\
    \cmidrule(lr){4-6} \cmidrule(lr){7-11} \cmidrule(lr){12-16}
     & & \tH{Diameter} & \tH{Visibility} & \tH{Pixels} & \tH{Distance} & \tH{Median} & \multicolumn{2}{c}{\tH{Precision (\%)}} & \multicolumn{2}{c}{\tH{Recall (\%)}} & \tH{Median} & \multicolumn{2}{c}{\tH{Precision (\%)}} & \multicolumn{2}{c}{\tH{Recall (\%)}} \\
     & \tH{Object} & \tH{(cm)} & \tH{(\%)} & \tH{($\times10^3$)} & \tH{(cm)} & \tH{Error (cm)} & \tH{@2cm} & \tH{@10cm} & \tH{@2cm} & \tH{@10cm} & \tH{Error (cm)} & \tH{@2cm} & \tH{@10cm} & \tH{@2cm} & \tH{@10cm} \\

    \midrule

    \ccw &         Mushrooms &   7.6 &  77.7 &  13.5 &  68.5 &   0.3 &  87.9 & 100.0 &  82.9 &  94.3 &   0.3 &  87.6 & 100.0 &  61.4 &  70.1 \\
    \ccw &              Tuna &   7.6 &  82.0 &  12.3 &  73.0 &   0.4 & 100.0 & 100.0 &  88.6 &  88.6 &   0.3 &  90.9 &  98.7 &  63.1 &  68.5 \\
    \ccw &            Yogurt &   8.3 &  88.1 &  18.6 &  72.3 &   0.3 &  94.1 & 100.0 &  80.0 &  85.0 &   0.3 &  98.8 & 100.0 &  83.2 &  84.2 \\
    \ccw &           Peaches &   8.9 &  86.4 &  21.9 &  74.4 &   0.2 & 100.0 & 100.0 &  85.0 &  85.0 &   0.3 &  80.8 & 100.0 &  64.6 &  80.0 \\
    \ccw &         Pineapple &   8.9 &  82.5 &  18.5 &  70.5 &   0.3 &  97.3 & 100.0 &  65.5 &  67.3 &   0.3 &  96.4 & 100.0 &  79.0 &  82.0 \\
    \ccw &          Cherries &   9.0 &  87.5 &  22.5 &  67.5 &   0.4 &  73.3 & 100.0 &  55.0 &  75.0 &   0.3 &  87.4 &  93.7 &  79.9 &  85.6 \\
    \ccw &              Corn &   9.0 &  81.1 &  20.6 &  69.4 &   0.3 & 100.0 & 100.0 &  57.5 &  57.5 &   0.3 &  91.1 & 100.0 &  67.8 &  74.4 \\
    \ccw &       Green Beans &   9.0 &  86.3 &  22.5 &  69.9 &   0.2 &  93.1 & 100.0 &  77.1 &  82.9 &   0.3 &  92.5 &  97.2 &  76.7 &  80.6 \\
    \ccw &   Peas \& Carrots &   9.0 &  88.1 &  21.7 &  70.3 &   0.2 & 100.0 & 100.0 &  73.3 &  73.3 &   0.3 &  99.0 & 100.0 &  81.6 &  82.4 \\
    \ccw &      Tomato Sauce &  10.7 &  83.6 &  17.9 &  79.9 &   0.2 & 100.0 & 100.0 &  80.0 &  80.0 &   0.3 & 100.0 & 100.0 &  76.4 &  76.4 \\
    \ccw &     Alphabet Soup &  10.8 &  85.0 &  27.1 &  70.8 &   0.2 &  96.9 & 100.0 &  68.9 &  71.1 &   0.3 &  93.5 &  98.9 &  78.9 &  83.5 \\
    \ccw &          Parmesan &  12.3 &  85.5 &  26.1 &  77.4 &   0.3 & 100.0 & 100.0 &  83.3 &  83.3 &   0.4 & 100.0 & 100.0 &  97.8 &  97.8 \\
    \enspace\multirow{-13}{*}{\sc\begin{sideways}cylinders\end{sideways}}
    \ccw & \bfseries Mean & \bfseries   9.2 & \bfseries  84.5 & \bfseries  20.3 & \bfseries  72.0 & \bfseries   0.3 & \bfseries  95.2 & \bfseries 100.0 & \bfseries  74.8 & \bfseries  78.6 & \bfseries   0.3 & \bfseries  93.2 & \bfseries  99.0 & \bfseries  75.9 & \bfseries  80.4 \\

    \midrule

    \ccw & Chocolate Pudding &   9.8 &  88.4 &  17.4 &  70.9 &   0.3 &  96.2 & 100.0 &  83.3 &  86.7 &   0.3 &  77.8 & 100.0 &  69.5 &  89.4 \\
    \ccw &            Butter &  11.5 &  82.7 &  16.7 &  72.5 &   0.2 &  97.4 & 100.0 &  84.4 &  86.7 &   0.3 &  83.2 & 100.0 &  52.3 &  62.9 \\
    \ccw &      Cream Cheese &  11.6 &  85.0 &  17.9 &  76.6 &   0.2 & 100.0 & 100.0 & 100.0 & 100.0 &   0.3 &  78.3 & 100.0 &  66.4 &  84.8 \\
    \ccw &           Raisins &  15.1 &  76.9 &  31.0 &  73.8 &   0.4 &  92.9 & 100.0 &  65.0 &  70.0 &   0.4 &  96.5 &  98.8 &  69.7 &  71.4 \\
    \ccw &           Popcorn &  15.2 &  78.3 &  32.7 &  74.8 &   2.7 &  44.4 & 100.0 &  20.0 &  45.0 &   1.3 &  45.7 &  83.5 &  42.6 &  77.9 \\
    \ccw &              Milk &  20.5 &  82.5 &  45.1 &  80.5 &   0.7 &  75.0 & 100.0 &  72.0 &  96.0 &   4.7 &  45.0 &  99.2 &  40.9 &  90.2 \\
    \ccw &      Orange Juice &  20.6 &  78.1 &  50.4 &  75.3 &   0.8 &  60.0 &  84.0 &  60.0 &  84.0 &   4.9 &  44.1 & 100.0 &  43.8 &  99.3 \\
    \ccw &      Granola Bars &  20.6 &  77.3 &  47.8 &  78.2 &   0.8 &  43.5 &  78.3 &  40.0 &  72.0 &   0.6 &  79.1 &  91.8 &  66.9 &  77.7 \\
    \ccw &     Mac \& Cheese &  20.7 &  75.3 &  53.2 &  75.1 &   0.3 &  92.7 &  95.1 &  76.0 &  78.0 &   0.8 &  84.4 & 100.0 &  69.9 &  82.8 \\
    \ccw &           Cookies &  20.9 &  77.1 &  44.4 &  81.6 &   0.6 &  90.9 &  95.5 &  80.0 &  84.0 &   0.4 &  87.1 &  98.8 &  71.8 &  81.6 \\
    \ccw &         Spaghetti &  25.3 &  78.9 &  35.2 &  74.5 &   0.6 &  85.3 & 100.0 &  82.9 &  97.1 &   1.1 &  74.3 &  98.2 &  63.6 &  84.1 \\
    \enspace\multirow{-12}{*}{\sc\begin{sideways}cuboids\end{sideways}}
    \ccw & \bfseries Mean & \bfseries  17.4 & \bfseries  80.0 & \bfseries  35.6 & \bfseries  75.8 & \bfseries   0.7 & \bfseries  79.8 & \bfseries  95.7 & \bfseries  69.4 & \bfseries  81.8 & \bfseries   1.4 & \bfseries  72.3 & \bfseries  97.3 & \bfseries  59.8 & \bfseries  82.0 \\

    \midrule

    \ccw &    Salad Dressing &  15.1 &  89.0 &  27.7 &  72.9 &   0.3 & 100.0 & 100.0 & 100.0 & 100.0 &   0.4 &  96.2 & 100.0 &  90.0 &  93.6 \\
    \ccw &              Mayo &  15.3 &  87.3 &  28.5 &  77.0 &   0.3 & 100.0 & 100.0 & 100.0 & 100.0 &   0.4 &  97.4 & 100.0 &  88.9 &  91.3 \\
    \ccw &         BBQ Sauce &  15.4 &  87.1 &  27.1 &  73.7 &   0.4 & 100.0 & 100.0 &  93.3 &  93.3 &   0.5 &  98.0 &  99.0 &  93.2 &  94.2 \\
    \ccw &           Ketchup &  15.4 &  89.8 &  29.5 &  72.8 &   0.3 &  94.7 & 100.0 &  90.0 &  95.0 &   0.3 &  98.4 &  98.4 &  83.8 &  83.8 \\
    \ccw &           Mustard &  16.1 &  89.2 &  31.5 &  71.8 &   0.5 & 100.0 & 100.0 &  96.0 &  96.0 &   0.5 &  95.6 & 100.0 &  92.1 &  96.4 \\
    \enspace\multirow{-6}{*}{\sc\begin{sideways}bottles\end{sideways}}
    \ccw & \bfseries Mean & \bfseries  15.5 & \bfseries  88.5 & \bfseries  28.9 & \bfseries  73.6 & \bfseries   0.4 & \bfseries  98.9 & \bfseries 100.0 & \bfseries  95.9 & \bfseries  96.9 & \bfseries   0.4 & \bfseries  97.1 & \bfseries  99.5 & \bfseries  89.6 & \bfseries  91.8 \\

    \midrule

    \ccw & \bfseries Overall Mean & \bfseries  13.6 & \bfseries  83.4 & \bfseries  27.8 & \bfseries  73.8 & \bfseries   0.5 & \bfseries  89.8 & \bfseries  98.3 & \bfseries  76.4 & \bfseries  83.1 & \bfseries   0.7 & \bfseries  85.7 & \bfseries  98.4 & \bfseries  72.0 & \bfseries  83.1 \\

    \bottomrule

\end{tabular}
\endgroup
}
\end{table*}

\section{Relationship to Previous Work}

Many existing datasets focus on the task of
6-DoF pose estimation of known objects \cite{xiang2018rss:posecnn,tremblay2018arx:fat,hinterstoisser2012accv:linemod,hodan2017wacv:tless,hodan2018eccv:bop,
tang2019cvpr:cityflow,xiang2016eccv:objectnet3d,xiang2004wacv}.
The original LineMOD dataset \cite{hinterstoisser2012accv:linemod},
for example, offers manual object
annotations for approximately 1000 images for each of the 15 objects in the dataset.
The LineMOD-Occluded dataset consists of additional annotations of the original dataset \cite{brachmann2014eccv:occlusion}.
T-LESS consists of real images of untextured manufacturing parts \cite{hodan2017wacv:tless}.
The falling things (FAT) dataset~\cite{tremblay2018arx:fat} consists of synthetically generated images of
random YCB \cite{calli2015ram:ycb} objects which fall under the influence of simulated gravity in 3D scenes.
The YCB-Video dataset provides a large number of images from video sequences, with high correlation between images and an average of five objects
visible per image \cite{xiang2018rss:posecnn,calli2015icar:ycb}.
The YCBInEOAT (``in the end of arm tooling'') dataset~\cite{bowen2020iros:se3tracknet} consists of videos of 5 YCB objects, one at a time, being held and moved.
The DexYCB dataset~\cite{yuwei2021cvpr:dexycb} contains multi-camera videos of humans grasping 20 YCB objects, one at a time.
Our HOPE dataset contains 28 objects in 50 cluttered scenes, nearly 5 lighting variations per scene (for a total of 238 images), and an average of more than 18 objects per scene.
The HOPE objects also appear in our HOPE-Video dataset~\cite{lin2021fusion:multilevel}, a collection of 10 short RGBD video sequences captured by a robot arm-mounted camera, each depicting 5-20 objects on a tabletop workspace.

Recently the BOP (Benchmark for 6D Object Pose Estimation) challenge~\cite{hodan2018eccv:bop} proposed assembling multiple 6-DoF
pose estimation datasets as a central resource for evaluating algorithms.
It is composed of the following datasets:
LineMOD \cite{hinterstoisser2012pami:linemod},
LineMOD-Occluded \cite{brachmann2014eccv:occlusion},
T-LESS \cite{hodan2017wacv:tless},
ITODD \cite{drost2017iccvw:pose},
HomebrewedDB \cite{kaskman2019arx:homebrew},
YCB-Video \cite{xiang2018rss:posecnn},
Rutgers APC \cite{rennie2016ral:rutgersapc},
IC-BIN \cite{doumanoglou2016cvpr:nextbestview},
IC-MI \cite{tejani2014eccv:3dobj},
TUD Light \cite{hodan2018eccv:bop},
Toyota Light \cite{hodan2018eccv:bop},
and now HOPE.
With the addition of these HOPE objects, we believe that this benchmark is more relevant to robotics researchers.

\section{Conclusion}
In this work we offer a pose estimation benchmark that is immediately applicable to robotic manipulation research.
Researchers can render synthetic images using the 3D textured meshes accompanying our dataset to train pose estimation models for 28 different objects. These models can be evaluated using our challenging, accurately annotated real-world images. Finally, pose predictions can be used directly in robotics labs for detecting and manipulating physical copies of the same objects, which are readily available from online retailers.
We hope that this dataset will be a useful and practical bridge between researchers in computer vision and robotics.

\section*{Acknowledgements}
The authors would like to thank Tom{\'a}{\v{s}} Hoda{\v{n}} for his feedback and assistance throughout this work and for integrating the HOPE dataset into the BOP benchmark.

\bibliographystyle{IEEEtran}
\bibliography{main}

\ifarxiv
\vfill

\pagebreak

\onecolumn

\section*{Appendix}

\subsection{HOPE Object Set}
The 28 toy grocery objects come from the following products, with links for purchasing:
\begin{itemize}
    \item[]
	\item \textbf{Melissa \& Doug Let's Play House! Grocery Shelf Boxes}
    \item[] ``Chocolate Pudding'', ``Cookies'', ``Granola Bars'', ``Mac \& Cheese'', ``Popcorn'', ``Raisins'', and ``Spaghetti''
    \item[] Links: \href{https://www.melissaanddoug.com/lets-play-house-grocery-shelf-boxes/5501.html}{Melissa \& Doug}, 
    \href{https://www.amazon.com/gp/product/B071ZMT9S2}{Amazon}, 
    \href{https://www.google.com/shopping/product/9344695624776110717}{Google Shopping}
	\item[]

	\item \textbf{Melissa \& Doug Let's Play House! Grocery Cans}
    \item[] ``Alphabet Soup'', ``Cherries'', ``Corn'', ``Green Beans'', ``Mushrooms'', ``Peaches'', ``Peas \& Carrots'', ``Pineapple'', ``Tomato Sauce'', and ``Tuna''
    \item[] Links: \href{https://www.melissaanddoug.com/lets-play-house-grocery-cans/4088.html}{Melissa \& Doug}, \href{https://www.amazon.com/gp/product/B007EA6PKS}{Amazon}, \href{https://www.google.com/shopping/product/3614702276975986631}{Google Shopping}
    \item[]
	
	\item \textbf{Melissa \& Doug Let's Play House! Fridge Fillers}
    \item[] ``Butter'', ``Cream Cheese'', ``Milk'', ``Orange Juice'', ``Parmesan'', and ``Yogurt''
    \item[] Links: \href{https://www.melissaanddoug.com/lets-play-house-fridge-fillers/4316.html}{Melissa \& Doug}, 
    \href{https://www.amazon.com/gp/product/B00H4SKSPS}{Amazon}, 
    \href{https://www.google.com/shopping/product/11269590935366386681}{Google Shopping}
    \item[]
	
	\item \textbf{Melissa \& Doug Favorite Condiments}
    \item[] ``BBQ Sauce'', ``Ketchup'', ``Mayo'', ``Mustard'', and ``Salad Dressing''
    \item[] Links: \href{https://www.melissaanddoug.com/favorite-condiments/4317.html}{Melissa \& Doug}, 
    \href{https://www.amazon.com/gp/product/B072M2PGX9}{Amazon}, 
    \href{https://www.amazon.com/gp/product/B072M2PGX9}{Google Shopping}
    \item[]

\end{itemize}
Note that some items included in these products (\emph{viz.,} ``Crackers'', ``Crispy Crisps'', ``Pancake Mix'', ``Deli Cheese Slices'', and ``Deli Slice Meats'') were omitted from our dataset because their sizes made scanning difficult.

\subsection{Results with ADD-H Metric}

Here we include versions of both Fig.~\ref{fig:meanssdresults}\,(a-c) and Table~\ref{tab:meanssdresults} using the ADD-H metric instead of MeanSSD.
\begin{figure*}[h]
    \setlength{\tabcolsep}{0pt}
    \centering
    \begin{tabular}{ccc}
        \multicolumn{3}{c}{\scriptsize\sffamily \textbf{Validation Set}} \\
        \begin{subfigure}[b]{0.24\linewidth}
            \centering\includegraphics[trim=4px 4px 4px 4px, clip, width=\linewidth]{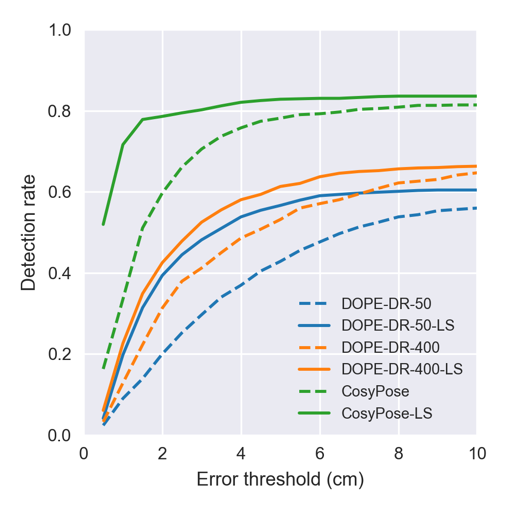}
        \end{subfigure} &
        \begin{subfigure}[b]{0.24\linewidth}
            \centering\includegraphics[trim=4px 4px 4px 4px, clip, width=\linewidth]{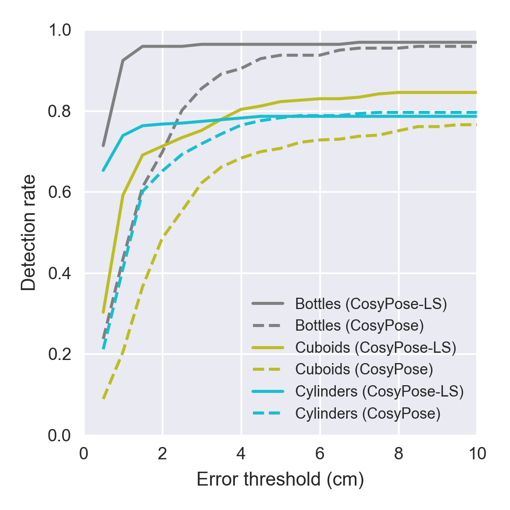}
        \end{subfigure} &
        \begin{subfigure}[b]{0.24\linewidth}
            \centering\includegraphics[trim=4px 4px 4px 4px, clip, width=\linewidth]{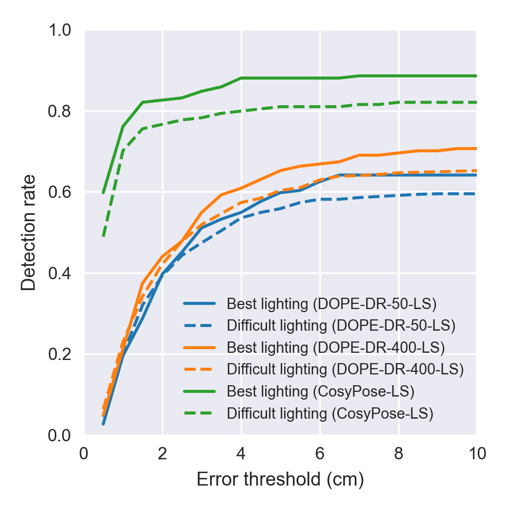}
        \end{subfigure} \\
        
        \multicolumn{3}{c}{\scriptsize\sffamily \textbf{Test Set}} \\
        \begin{subfigure}[b]{0.24\linewidth}
            \centering\includegraphics[trim=4px 4px 4px 4px, clip, width=\linewidth]{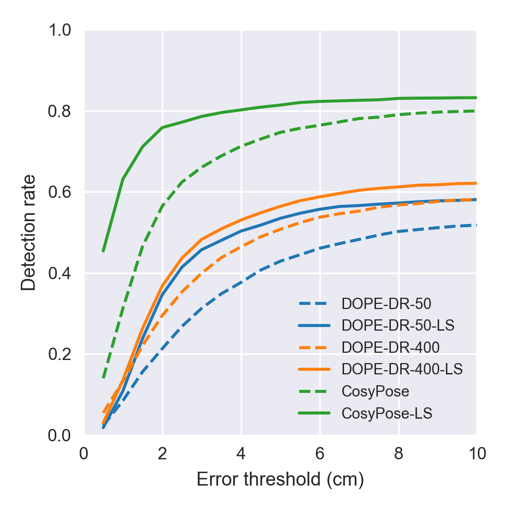}
        \end{subfigure} &
        \begin{subfigure}[b]{0.24\linewidth}
            \centering\includegraphics[trim=4px 4px 4px 4px, clip, width=\linewidth]{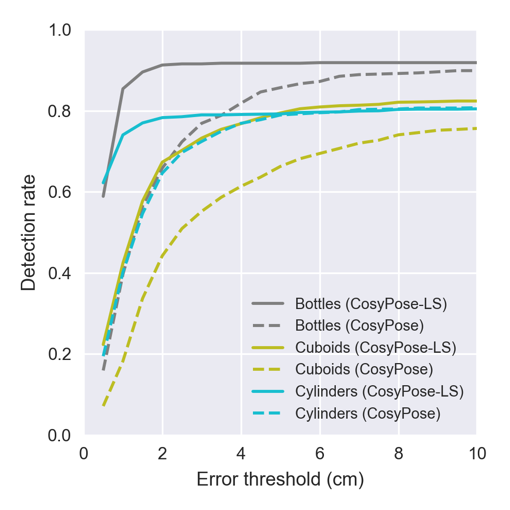}
        \end{subfigure} &
        \begin{subfigure}[b]{0.24\linewidth}
            \centering\includegraphics[trim=4px 4px 4px 4px, clip, width=\linewidth]{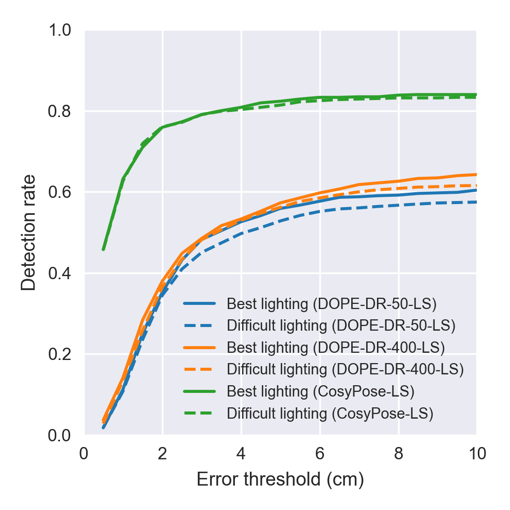}
        \end{subfigure} \\
        
        \vspace{-0.25em}
        \footnotesize{(a) Methods} &
        \footnotesize{(b) Object categories} &
        \footnotesize{(c) Lighting} \\
        
        \scriptsize{(ADD-H)} &
        \scriptsize{(ADD-H + CosyPose-LS)} &
        \scriptsize{(ADD-H)} \\
    \end{tabular}
    \caption{Detection rate versus maximum detection threshold, computed on the HOPE validation set (top row) and test set (bottom row).}
    \label{fig:addhresults}
\end{figure*}

\begin{table*}[t]
    \centering
    \caption{Results on the validation and test sets using CosyPose with the ADD-H metric. (The median error was computed among all ``true positive'' predictions, determined at a 10~cm error threshold.)
    }
    \label{tab:addhresults}
    \resizebox{0.97\textwidth}{!}{%
    \begingroup
    \setlength{\tabcolsep}{5pt} %
    \renewcommand{\arraystretch}{1.1} %
    \rowcolors{6}{gray!10}{white!0}
        \begin{tabular}{@{}rlcccccccccccccc} %
            
            \toprule
            & & & \multicolumn{3}{c}{\tH{Pose Statistics}} & \multicolumn{5}{c}{\tH{Validation Set}} & \multicolumn{5}{c}{\tH{Test Set}} \\
            \cmidrule(lr){4-6} \cmidrule(lr){7-11} \cmidrule(lr){12-16}
            & & \tH{Diameter} & \tH{Visibility} & \tH{Pixels} & \tH{Distance} & \tH{Median} & \multicolumn{2}{c}{\tH{Precision (\%)}} & \multicolumn{2}{c}{\tH{Recall (\%)}} & \tH{Median} & \multicolumn{2}{c}{\tH{Precision (\%)}} & \multicolumn{2}{c}{\tH{Recall (\%)}} \\
            & \tH{Object} & \tH{(cm)} & \tH{(\%)} & \tH{($\times10^3$)} & \tH{(cm)} & \tH{Error (cm)} & \tH{@2cm} & \tH{@10cm} & \tH{@2cm} & \tH{@10cm} & \tH{Error (cm)} & \tH{@2cm} & \tH{@10cm} & \tH{@2cm} & \tH{@10cm} \\
            
            \midrule
            
            \ccw &         Mushrooms &   7.6 &  77.7 &  13.5 &  68.5 &   0.3 & 100.0 & 100.0 &  94.3 &  94.3 &   0.3 &  97.8 & 100.0 &  68.5 &  70.1 \\
            \ccw &              Tuna &   7.6 &  84.4 &  12.7 &  70.9 &   0.3 & 100.0 & 100.0 &  88.6 &  88.6 &   0.3 &  92.2 &  98.7 &  64.0 &  68.5 \\
            \ccw &            Yogurt &   8.3 &  88.1 &  18.6 &  72.3 &   0.3 & 100.0 & 100.0 &  85.0 &  85.0 &   0.3 &  95.3 & 100.0 &  80.2 &  84.2 \\
            \ccw &           Peaches &   8.9 &  87.5 &  22.5 &  72.3 &   0.2 & 100.0 & 100.0 &  85.0 &  85.0 &   0.3 &  91.3 & 100.0 &  73.1 &  80.0 \\
            \ccw &         Pineapple &   8.9 &  82.6 &  18.5 &  70.5 &   0.3 & 100.0 & 100.0 &  67.3 &  67.3 &   0.3 &  98.5 & 100.0 &  80.8 &  82.0 \\
            \ccw &          Cherries &   9.0 &  87.7 &  22.5 &  67.5 &   0.4 &  73.3 & 100.0 &  55.0 &  75.0 &   0.3 &  89.8 &  93.7 &  82.0 &  85.6 \\
            \ccw &              Corn &   9.0 &  81.1 &  20.6 &  69.4 &   0.3 & 100.0 & 100.0 &  57.5 &  57.5 &   0.3 & 100.0 & 100.0 &  74.4 &  74.4 \\
            \ccw &       Green Beans &   9.0 &  86.3 &  22.5 &  69.9 &   0.2 &  96.6 & 100.0 &  80.0 &  82.9 &   0.3 &  94.4 &  97.2 &  78.3 &  80.6 \\
            \ccw &   Peas \& Carrots &   9.0 &  88.1 &  21.7 &  70.3 &   0.2 & 100.0 & 100.0 &  73.3 &  73.3 &   0.3 & 100.0 & 100.0 &  82.4 &  82.4 \\
            \ccw &      Tomato Sauce &  10.7 &  84.8 &  18.5 &  78.4 &   0.2 & 100.0 & 100.0 &  80.0 &  80.0 &   0.3 & 100.0 & 100.0 &  76.4 &  76.4 \\
            \ccw &     Alphabet Soup &  10.8 &  86.7 &  27.9 &  69.5 &   0.3 & 100.0 & 100.0 &  71.1 &  71.1 &   0.3 &  96.7 &  98.9 &  81.7 &  83.5 \\
            \ccw &          Parmesan &  12.3 &  88.0 &  28.0 &  74.1 &   0.4 & 100.0 & 100.0 &  83.3 &  83.3 &   0.4 & 100.0 & 100.0 &  97.8 &  97.8 \\
            \enspace\multirow{-13}{*}{\sc\begin{sideways}cylinders\end{sideways}}
            \ccw & \bfseries Mean & \bfseries   9.2 & \bfseries  85.2 & \bfseries  20.6 & \bfseries  71.1 & \bfseries   0.3 & \bfseries  97.5 & \bfseries 100.0 & \bfseries  76.7 & \bfseries  78.6 & \bfseries   0.3 & \bfseries  96.3 & \bfseries  99.0 & \bfseries  78.3 & \bfseries  80.4 \\
            
            \midrule
            
            \ccw & Chocolate Pudding &   9.8 &  88.4 &  17.4 &  70.9 &   0.3 &  96.2 & 100.0 &  83.3 &  86.7 &   0.3 &  82.2 & 100.0 &  73.5 &  89.4 \\
            \ccw &            Butter &  11.5 &  82.7 &  16.7 &  72.5 &   0.2 &  97.4 & 100.0 &  84.4 &  86.7 &   0.6 &  84.2 & 100.0 &  53.0 &  62.9 \\
            \ccw &      Cream Cheese &  11.6 &  87.9 &  18.5 &  75.1 &   0.8 & 100.0 & 100.0 & 100.0 & 100.0 &   0.8 &  84.9 & 100.0 &  72.0 &  84.8 \\
            \ccw &           Raisins &  15.1 &  78.7 &  31.0 &  73.8 &   0.9 &  92.9 & 100.0 &  65.0 &  70.0 &   1.0 &  96.5 &  98.8 &  69.7 &  71.4 \\
            \ccw &           Popcorn &  15.2 &  78.3 &  32.7 &  74.8 &   2.5 &  44.4 & 100.0 &  20.0 &  45.0 &   1.5 &  46.5 &  83.5 &  43.4 &  77.9 \\
            \ccw &              Milk &  20.5 &  82.3 &  46.3 &  79.2 &   0.7 &  87.5 & 100.0 &  84.0 &  96.0 &   0.9 &  87.5 & 100.0 &  79.5 &  90.9 \\
            \ccw &      Orange Juice &  20.6 &  78.1 &  50.4 &  75.3 &   0.8 &  68.0 &  92.0 &  68.0 &  92.0 &   1.3 &  80.0 &  99.3 &  79.5 &  98.6 \\
            \ccw &      Granola Bars &  20.6 &  79.0 &  49.2 &  76.6 &   3.2 &  43.5 & 100.0 &  40.0 &  92.0 &   1.6 &  74.5 &  95.5 &  63.1 &  80.8 \\
            \ccw &     Mac \& Cheese &  20.7 &  75.3 &  53.2 &  75.0 &   1.1 &  92.7 &  97.6 &  76.0 &  80.0 &   1.2 &  85.7 & 100.0 &  71.0 &  82.8 \\
            \ccw &           Cookies &  20.9 &  77.1 &  44.4 &  81.6 &   1.3 &  90.9 &  95.5 &  80.0 &  84.0 &   1.2 &  88.2 & 100.0 &  72.8 &  82.5 \\
            \ccw &         Spaghetti &  25.3 &  78.9 &  36.1 &  73.3 &   0.6 &  85.3 & 100.0 &  82.9 &  97.1 &   1.2 &  73.5 &  98.2 &  62.9 &  84.1 \\
            \enspace\multirow{-12}{*}{\sc\begin{sideways}cuboids\end{sideways}}
            \ccw & \bfseries Mean & \bfseries  17.4 & \bfseries  80.6 & \bfseries  36.0 & \bfseries  75.3 & \bfseries   1.1 & \bfseries  81.7 & \bfseries  98.6 & \bfseries  71.2 & \bfseries  84.5 & \bfseries   1.0 & \bfseries  80.3 & \bfseries  97.8 & \bfseries  67.3 & \bfseries  82.4 \\
            
            \midrule
            
            \ccw &    Salad Dressing &  15.1 &  89.0 &  27.7 &  72.9 &   0.3 & 100.0 & 100.0 & 100.0 & 100.0 &   0.4 & 100.0 & 100.0 &  93.6 &  93.6 \\
            \ccw &              Mayo &  15.3 &  87.4 &  28.5 &  76.9 &   0.3 & 100.0 & 100.0 & 100.0 & 100.0 &   0.4 &  99.1 & 100.0 &  90.5 &  91.3 \\
            \ccw &         BBQ Sauce &  15.4 &  88.0 &  27.1 &  73.7 &   0.4 & 100.0 & 100.0 &  93.3 &  93.3 &   0.5 &  99.0 &  99.0 &  94.2 &  94.2 \\
            \ccw &           Ketchup &  15.4 &  90.2 &  29.5 &  72.8 &   0.3 &  94.7 & 100.0 &  90.0 &  95.0 &   0.3 &  98.4 &  98.4 &  83.8 &  83.8 \\
            \ccw &           Mustard &  16.1 &  89.2 &  31.5 &  71.7 &   0.5 & 100.0 & 100.0 &  96.0 &  96.0 &   0.5 &  97.8 & 100.0 &  94.3 &  96.4 \\
            \enspace\multirow{-6}{*}{\sc\begin{sideways}bottles\end{sideways}}
            \ccw & \bfseries Mean & \bfseries  15.5 & \bfseries  88.8 & \bfseries  28.9 & \bfseries  73.6 & \bfseries   0.4 & \bfseries  98.9 & \bfseries 100.0 & \bfseries  95.9 & \bfseries  96.9 & \bfseries   0.4 & \bfseries  98.9 & \bfseries  99.5 & \bfseries  91.3 & \bfseries  91.8 \\
            
            \midrule
            
            \ccw & \bfseries Overall Mean & \bfseries  13.6 & \bfseries  84.0 & \bfseries  28.1 & \bfseries  73.2 & \bfseries   0.6 & \bfseries  91.5 & \bfseries  99.5 & \bfseries  78.0 & \bfseries  84.2 & \bfseries   0.6 & \bfseries  90.5 & \bfseries  98.6 & \bfseries  76.3 & \bfseries  83.2 \\
            
            \bottomrule
            
        \end{tabular}
        \endgroup
        }
    \end{table*}
    \fi
    
    \end{document}